\documentclass[11pt,a4paper]{article}
\usepackage[hyperref]{acl2021}
\usepackage{times,latexsym}

\usepackage{url}
\usepackage[T1]{fontenc}
\usepackage{microtype}

\aclfinalcopy 
\def\aclpaperid{2423} 

\usepackage{xspace,mfirstuc,tabulary}

\usepackage{xspace,mfirstuc,tabulary}
\usepackage{amsmath}
\usepackage{multirow}
\usepackage{comment}
\usepackage{xcolor}
\usepackage{ulem}
\usepackage{paralist}
\usepackage{arydshln}

\usepackage[utf8]{inputenc}
\usepackage{pgfplots}
\pgfplotsset{compat=newest}
\usepgfplotslibrary{groupplots}
\usepgfplotslibrary{dateplot}

\makeatletter
\newcommand{\@emptybiblabel}[1]{}
\makeatother
\usepackage{color}

\usepackage{xcolor}
\definecolor{gblue}{RGB}{66,133,244}
\definecolor{gred}{RGB}{219,68,55}
\definecolor{gyellow}{RGB}{244,160,0}
\definecolor{ggreen}{RGB}{15,157,88}

\usepackage{algorithm}
\usepackage{algpseudocode}
\usepackage{bold-extra}
\usepackage{wrapfig}
\definecolor{deepcarminepink}{rgb}{0.94, 0.19, 0.22}
\definecolor{azure}{rgb}{0.0, 0.5, 1.0}

\algrenewcommand\alglinenumber[1]{\tiny #1:}
\renewcommand{\algorithmiccomment}[1]{{\color{blue}
\hfill$\triangleright${\textit{ #1}}}}

\newcommand\yumo[1]{\textcolor{red}{\textbf{#1}}}

\usepackage{amssymb}
\usepackage{pifont}
\newcommand{\cmark}{\ding{51}}%
\newcommand{\xmark}{\ding{55}}%

\usepackage{helvet}  
\usepackage{courier}  
\usepackage{graphicx}  
\usepackage{amssymb}
\usepackage{amsfonts}
\usepackage{paralist}
\usepackage{bbold}
\usepackage{bm}
\usepackage[toc,page]{appendix} 
\usepackage{float}
\usepackage{makecell}
\usepackage{subfig}

\DeclareSymbolFontAlphabet{\amsmathbb}{AMSb}%

\DeclareMathOperator*{\concat}{concat}

\DeclareMathOperator*{\rougeone}{R_1}
\DeclareMathOperator*{\rougetwo}{R_2}

\newcommand{\indep}{\perp \!\!\! \perp}

\usepackage{pict2e,picture}
\makeatletter
\DeclareRobustCommand{\Arrow}[1][]{%
\check@mathfonts
\if\relax\detokenize{#1}\relax
\settowidth{\dimen@}{$\m@th\rightarrow$}%
\else
\setlength{\dimen@}{#1}%
\fi
\sbox\z@{\usefont{U}{lasy}{m}{n}\symbol{41}}%
\begin{picture}(\dimen@,\ht\z@)
\roundcap
\put(\dimexpr\dimen@-.7\wd\z@,0){\usebox\z@}
\put(0,\fontdimen22\textfont2){\line(1,0){\dimen@}}
\end{picture}%
}
\makeatother

\usepackage{stmaryrd}
\usepackage{bibentry}
\usepackage{commath}

\begin{document}
\title{Generating Query Focused Summaries from Query-Free Resources}
 \author{Yumo Xu \and Mirella Lapata\\
 Institute for Language, Cognition and Computation\\
 School of Informatics, University of Edinburgh\\
 10 Crichton Street, Edinburgh EH8 9AB\\
 \texttt{yumo.xu@ed.ac.uk} \quad
 \texttt{mlap@inf.ed.ac.uk}}

\date{}

\makeatletter
\newcommand{\thickhline}{%
    \noalign {\ifnum 0=`}\fi \hrule height 1pt
    \futurelet \reserved@a \@xhline
}
\makeatother

\maketitle
\begin{abstract}
The availability of large-scale datasets has driven the development of
neural models that create \emph{generic} summaries from single or
multiple documents. In this work we consider \emph{query focused}
summarization (QFS), a task for which training data in the form of
queries, documents, and summaries is not readily available. We propose
to decompose QFS into (1)~query modeling (i.e.,~finding supportive
evidence within a set of documents for a query) and (2)~conditional
language modeling (i.e.,~summary generation). We introduce
\textsc{Marge}, a \textbf{Ma}sked \textbf{R}OU\textbf{GE} Regression
framework for evidence estimation and ranking which relies on a
unified representation for summaries and queries, so that summaries in
generic data can be converted into \emph{proxy queries} for learning a
query model. Experiments across QFS benchmarks and query types show
that our model achieves state-of-the-art performance despite learning from weak
supervision.\footnote{Our code and data is available at \url{https://github.com/yumoxu/marge}.}
\end{abstract}

\vspace*{-.2cm}
\begin{figure*}[t]
\includegraphics[width=16cm]{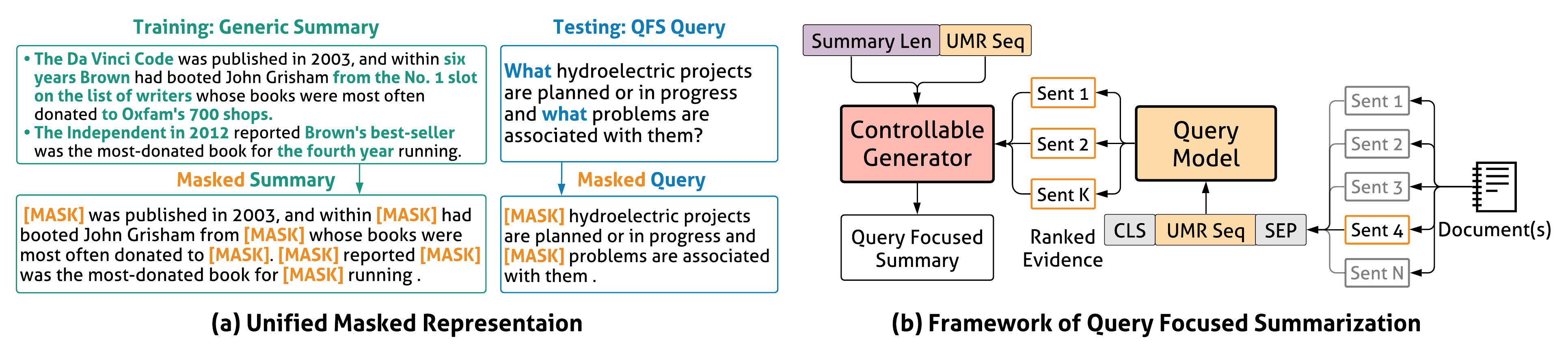}
\vspace*{-.7cm}
	\caption{\label{fig:frame}
          Overview of our abstractive QFS approach.  Summaries and
          queries are rendered with Unified Masked Representation
          (UMR) for training and testing, respectively.  The
          summarization framework consists of a query model and a
          controllable generator.  The query model ranks sentences in
          the input document(s) which provide evidence to answer the
          query; the generator operates over evidence bearing
          sentences to generate the final summary.}
\end{figure*}


\section{Introduction}
\label{sec:intro}
The neural encoder-decoder framework has become increasingly popular
in \emph{generic} summarization
(\citealt{see2017get,gehrmann2018bottom,liu2019hierarchical,fabbri2019multi},
\emph{inter alia}) thanks to the availability of large-scale datasets
containing hundreds of thousands of document-summary pairs. Training
data of this magnitude is not readily available for \emph{query
focused} summarization (QFS; \citealt{dang2005overview}) which aims to
create a short summary from a set of documents that answers a specific
query.  Existing corpora
\cite{nema2017diversity,dang2005overview,hoa2006overview,baumel2016topic}
are relatively small for modern data-hungry neural architectures and
have been mostly used for evaluation purposes.

A major bottleneck in leveraging generic summarization data for QFS is
the absence of queries \cite{nema2017diversity}; 
the majority of existing datasets consist of document-summary pairs, while QFS summaries are
expected to answer specific queries. Recent work
\cite{xu2020query,su2020caire,laskar2020wsl} sidesteps this problem by
resorting to distant supervision from query-relevant NLP resources
including question answering
\cite{rajpurkar2016squad,chakraborty2020biomedbert} and paraphrase
identification \cite{mrpc}. Such approaches incorporate query modeling
in the summarization process but are even more data hungry compared to
generic summarization ones, since they additionally require access to
QA datasets which can be extremely costly to create
\cite{bajaj2016ms,kwiatkowski2019natural}.  Moreover, there is often a
mismatch between queries in QA datasets and those in QFS scenarios
\cite{xu2020query}; the two types of queries are not identically
distributed and it is practically infeasible to find appropriate
query-related resources for all domains and topics.




In this work we do not assume access to any resources other than those
available for generic summarization. We further decompose abstractive
QFS into two subtasks: (1)~\textit{query modeling} (i.e.,~finding
supportive evidence within a set of documents for a query) and
(2)~\textit{conditional language modeling} (i.e., generating an
abstractive summary based on found evidence).  Under this formulation,
we use generic summarization data not only for conditional language
modeling, but also for learning an evidence ranking model.
Inspired by the Cloze task and its applications in NLP
\cite{taylor1953cloze,lewis2019unsupervised,lee2019latentqa}, we
propose \textsc{Marge}, a \textbf{Ma}sked \textbf{R}OU\textbf{GE}
regression framework for evidence estimation and ranking.
\textsc{Marge} introduces a unified representation for
\textit{summaries} and \textit{queries}, so that summaries in generic
data can be converted into \textit{proxy queries} for learning a query
model.
Based on the evidence selected by \textsc{Marge}, we
generate abstractive summaries  whilst controlling their length and the
extent to which the query influences their content. 


Our contributions in this work are threefold: we propose a weakly
supervised system for abstractive QFS where no query-related resources
are required; we discover a new type of connection between generic
summaries and QFS queries, and provide a universal representation for
them which allows generic summarization data to be exploited
for QFS; we provide experimental results on QFS benchmarks, and show
that across query types and domains our system achieves
state-of-the-art results on both evidence ranking and abstractive QFS.

\section{Related Work}
\label{sec:related}

The majority of previous QFS approaches have been extractive, operating over
queries and document clusters from which they select query-relevant
sentences to compose a summary.  They mostly differ in the way
centrality and relevance are estimated and incorporated, e.g., via
manifold ranking \cite{wan2007manifold}, using a look-ahead strategy
\cite{badrinath2011improving}, uncertainty prediction
\cite{wan2014ctsum}, or attention mechanisms \cite{li2017cascaded,li2017salience}.
More recently \newcite{xu2020query} propose a coarse-to-fine framework
that leverages distant supervision from question answering to extract
summary-worthy content.


Abstractive QFS has received significantly less attention. This is due
to generation models being particularly data-hungry
\cite{lebanoff2018adapting,liu2019hierarchical} and the scarcity of
QFS training data. The increasing availability of pretrained models
has prompted the development of pipeline-style frameworks for QFS
which use resources from a wider range of NLP tasks.  For example,
\newcite{su2020caire} fine-tune BART \cite{bart} on CNN/DailyMail
\cite{hermann2015teaching}, a single-document summarization dataset,
and generate abstracts for QFS by iteratively summarizing paragraphs
to a budget.  They learn a query model for paragraph selection based
on a plethora of QA and machine reading datasets
\cite{su2019generalizing,rajpurkar2016squad}.  Similarly,
\newcite{laskar2020wsl} fine-tune \textsc{BertSum} on CNN/DailyMail,
and propose a three-stage system which uses supervision from QFS data
(typically reserved for evaluation) and related QA and paraphrase
identification tasks.

We also focus on abstractive QFS, however, we do not assume access to
any additional training resources over and above generic summarization
datasets, even for query modeling.
Moreover, our system is able to generate long QFS abstracts \emph{all
at once}, instead of \emph{iteratively} creating bullet-style
summaries which often lack coherence.

\section{Problem Formulation}
\label{sec:formulation}

Let \mbox{$\{(S, \mathcal{D})\}$} denote a
generic summarization dataset where \mbox{$\mathcal{D} = \{d_1, d_2,
  \dots, d_M\}$} is a collection of documents with corresponding
summaries~$S$. $|\mathcal{D}|=1$ for single-document summarization
(SDS) and $|\mathcal{D}| > 1$ for multi-document summarization (MDS).
In QFS, a query~$Q$ additionally specifies an information request,
\mbox{$\{(S, \mathcal{D}, Q)\}$}.  It is often assumed (e.g.,~in DUC
benchmarks) that $Q$~consists of a short title (e.g., \textsl{Amnesty
  International}), and a query narrative which is longer and more
detailed (e.g., \textsl{What is the scope of operations of Amnesty
  International and what are the international reactions to its
  activities?}).


In this work, we propose to decompose QFS into two sub-tasks, namely
\textit{query modeling} and \textit{conditional language modeling}.
The query model $q_{\theta}(D|Q;\theta)$ estimates whether textual
units (e.g.,~sentences) within document cluster~$D$ are relevant to
query~$Q$, while $p_{\phi}(S|D,Q;\phi)$ generates summary~$S$
conditioned on evidence provided by the query model and (optionally)
the query itself (see Figure~\ref{fig:frame}(b) for an illustration).
When $S \indep Q$, we have a \textit{query-agnostic} conditional
language model $p_{\phi}(S|D;\phi)$.  Otherwise, the conditional
language model is \textit{query-guided}. Our query model is trained
with distant supervision derived from \textit{generic} summarization
data which is easier to obtain (e.g.,~from online sources) compared to
QA datasets which must be annotated from scratch (e.g., for different
types of questions and domains).  Although queries are not verbalized
in generic summarization, we hypothesize that the summaries themselves
constitute a response to \textit{latent} queries.



So, how can we reverse-engineer the queries from the summaries?
Inspired by the standard Cloze task \cite{taylor1953cloze} and its
recent variants \cite{lewis2019unsupervised,lee2019latentqa}, we
render queries and summaries in a \textit{Unified Masked
  Representation} (UMR) which enables summaries to serve as
\textit{proxy} queries for model training, as shown in
Figure~\ref{fig:frame}(a).  We further assume that the answer to these
queries can be found in sentences which form part of the document
collection~$\mathcal{D}$. Although we do not know for certain what
these sentences are we can assume that if they have a high ROUGE score
against the reference summary they are likely to contain an answer. We
therefore use ROUGE as a distant supervision signal, and train a model
that takes a query and document sentence as input and estimates their
relevance.  At inference time, we also render actual queries in UMR
and rank all sentences in the document collection with our trained
model. The most relevant sentences serve as input to a conditional
language model to generate query focused abstractive summaries.

\section{Query Modeling}
\label{sec:marge}
\algtext*{EndWhile}
\algtext*{EndIf}
\algtext*{EndFor}

\begin{algorithm}[t]
\scriptsize
\caption{\label{algo}Generate Masked Summary}%
\begin{algorithmic}[1]
\renewcommand{\algorithmiccomment}[1]{{\color{blue}
\hfill$\triangleright${\textit{ #1}}}}
\Function{MaskSummary}{$\mathcal S$, $\gamma$}  
\Comment{Summary sentences and mask ratio}
\State Parse each $s \in S$ with OpenIE to extract information slots $\mathcal I$
\State Reveal budget $B = |\mathcal{I}|*\gamma$  \Comment{Reveal information partially}
\State Initialize revealed word number $b=0$
\State Initialize masked summary $\mathcal M$ to $S$ and fill with \texttt{[MASK]} 
\State Initialize $\texttt{EOM}=\text{false}$ \Comment{End of Masking}
\While{true}
    \State $\mathcal{S}_{a}=$\Call{GetAvaiable}{$\mathcal S$}  
    \Comment{Sentences with masked slots}
    \For{$s\gets \mathcal{S}_a$}
        \State $b=b+$ \Call{Reveal}{$s$}
        \Comment{Reveal a randomly sampled slot and record its length, i.e., \#tokens}
        \If{$b \ge B$} $\texttt{EOM}=\text{true}$
        \EndIf
    \EndFor
    \If{$\texttt{EOM}$}  
    \Comment{Start post-process}
        \For{$m\gets \mathcal M$}
            \State \Call{Merge}{$m$}  \Comment{Merge adjacent \texttt{[MASK]}}
        \EndFor
        \State \Return $\mathcal{M}$
    \EndIf
\EndWhile
\EndFunction
\end{algorithmic}
\end{algorithm}

As explained earlier, we train a query model $q_{\theta}(D|Q;\theta)$
on summary-sentence pairs via distant supervision. We use a
summary-based proxy query \texttt{UMR}$_{S}$ during training and an
actual query \texttt{UMR}$_{Q}$ during testing. 
In the following, we first describe how UMRs are obtained and then discuss how the query model is trained.

\paragraph{Unified Masked Representation}
The intuition behind UMR is that a summary will encapsulate most
salient information a user needs, while a query typically covers only
a small fraction. We thus add one or more ``placeholders'' to the query
to represent missing information the user actually seeks.  We also
identify such information in generic summaries for selective masking,
to reduce the distributional shift during training.

The UMR for a summary is the concatenation of its sentential UMRs.  To
convert a sentence from natural language to UMR, we parse it with Open
Information Extraction (Open IE; \citealt{stanovsky2018supervised}) to
a set of propositions consisting of verbs and their arguments.  The
latter are considered candidate \textit{information slots} $\mathcal{I}$.  
We initialize Algorithm~\ref{algo}, by replacing all such slots with a
\texttt{[MASK]} token.  We subsequently sample and reveal a set of
slots subject to a budget constraint.
We define the budget as $B=\gamma * |\mathcal{I}|$ where $\gamma \in
[0, 1]$ modulates the proportion of tokens to be revealed within
$\mathcal{I}$ slots (and is optimized on the development set).
Finally, in order to keep the representation of \texttt{UMR}$_{S}$ and
$\texttt{UMR}_{Q}$ consistent (see next paragraph), we merge adjacent
\texttt{[MASK]} tokens to one \texttt{[MASK]} resulting in a partially
masked summary.


We mask QFS queries by considering their structure and lexical
makeup. Queries in DUC benchmarks often contain \emph{interrogative}
words (e.g.,~\textsl{how is A} and \textsl{what is B}) and
\emph{request} words (e.g., \textsl{describe A} and \textsl{tell
  me~B}). Following this observation, we manually collect a small set
of such query words and replace them with \texttt{[MASK]}. For queries
with a title and a narrative, we first mask the narrative and then
prepend ``\texttt{[MASK]} $\mathcal T$ \texttt{.}'', where
$\mathcal{T}$ is a sequence of title tokens. Figure~\ref{fig:frame}(a)
shows examples of a masked query and summary.

\paragraph{Evidence Ranking}
We represent  sentences in a document collection and UMR queries
with a pre-trained BERT model \cite{devlin2019bert}. Specifically, we
concatenate a UMR query and a candidate sentence to sequence
``\texttt{[CLS]} $\mathcal U$ \texttt{[SEP]} $\mathcal C$
\texttt{[SEP]}'' where $\mathcal U$ is a sequence of tokens within a
UMR query and $\mathcal C$ a sequence of tokens in a document sentence
(we pad each sequence in a minibatch of~$L$ tokens).  The
\texttt{[CLS]} vector serves as input to a single layer neural network
which estimates whether the sentence contains sufficient evidence to
answer the query (see Figure~\ref{fig:frame}(b) right). We use the
mean-square error to compute the loss and update the encoding
parameters in BERT via standard backpropagation:
\begin{equation}
    \label{eq:loss}
    \mathcal{L}(\theta) =
    \frac{1}{|\mathcal{D}|}\sum_{(S,C) \sim \mathcal{D}}
    \left[(y-\hat{y}(S,C;\theta))^2\right].
\end{equation}
where $S,C$ is a summary-sentence pair sampled from collection~$D$
and~$y$ the training signal. Recall the summary is rendered as
\texttt{UMR}$_{S}$.


Previous work \cite{liu2019hierarchical} has used ROUGE-2 as training
signal for paragraph ranking. However, sentences are significantly
shorter than paragraphs, and we observe a number of instances with a
ROUGE-2 score of 0.  We therefore perform label smoothing and
define~$y$ as the F1 interpolation of ROUGE-2 and \mbox{ROUGE-1}:
\mbox{$y=\rougetwo(S, C) + \lambda * \rougeone(S,C)$} where $\lambda$
is optimized on the development set.  At inference time, we use the
trained model to compute the affinity score between \texttt{UMR}$_{Q}$
and all candidate sentences in~$\mathcal{D}$ and rank them
accordingly.  The highest ranked sentences are deemed query-relevant
and passed on to our summary generation model.\footnote{The Cloze task
has been also employed in recent work in generic
summarization \cite{huang-etal-2020-knowledge}.  In comparison, we
address a different research question (i.e., query modeling
vs. summary evaluation) based on a different formulation (masked ROUGE
regression vs. multiple-choice QA).}

\paragraph{Query Narrative Expansion} 
In some cases queries may be relatively short and narratives absent.
This can be problematic for our setup since query proxies (in the form
of summaries) are typically long and detailed. For datasets with short
queries we automatically create query narratives in an unsupervised
fashion.  We employ LexRank \cite{erkan2004lexrank} to select a subset
of representative sentences under a word budget
and concatenate them to form narratives (which we append to the
original queries).

\section{Query Focused Generation}
\label{sec:gen}
We also leverage {generic} summarization datasets to fine-tune a
pretrained language model for abstractive QFS.  In experiments we
employ the publicly released \textsc{UniLMv2} \cite{bao2020unilmv2} to
instantiate the controllable generator shown in
Figure~\ref{fig:frame}(b), however any other language model could have
been used instead.  

With Transformer \cite{vaswani2017attention} as the backbone network,
\textsc{UniLMv2} is jointly pretrained for natural language
understanding and generation.
Specifically, a bidirectional model is employs an autoencoding
objective (AE; identical to \citealt{devlin2019bert}), while a
partially autoregressive (PAR) sequence-to-sequence model decomposes
the probability of masked tokens in  input sequence $x$ as:
\begin{equation}
	\begin{aligned}
p(x_{M} \mid x_{\backslash M}) 
=\prod_{i=1}^{|M|} \prod_{m \in M_{i}} p(x_{m} \mid x_{\backslash M_{\geq i}})
\end{aligned}
\end{equation}  
where $M$ is the uniformly-produced  factorization order. 
The masked position set $M_i$ at the $i$th factorization step can be either a token or a $n$-gram block.
{$x_M$ is a set of $x_{M_i}$, and similarly, $x_{\backslash M}$ is a set of $x_{\backslash M_i}$.}
The pretraining loss is computed as $\mathcal{L}_{\text{AE}}+\mathcal{L}_{\text{PAR}}$.

At inference, \textsc{UniLMv2} operates over sentences deemed relevant
by the query model and decodes summaries autoregressively (see
Figure~\ref{fig:frame}(b) left).



\paragraph{Synthetic MDS Data}

The pre-trained language model can be fine-tuned on MDS datasets
(e.g.,~Multi-News; \citealt{fabbri2019multi}) which are perhaps better
aligned with the QFS task since both MDS and QFS operate over document
clusters. We additionally propose a way to create synthetic MDS
datasets based on SDS data. This is advantageous for two reasons.
Firstly, MDS resources are fairly limited compared to SDS data
\cite{zhang2018adapting,lebanoff2018adapting}. And secondly, by
construction, we can ensure various data characteristics which might
be desirable (e.g., the number of topics represented in the document
collection).


A challenge with leveraging SDS for QFS is the summary length
\cite{lebanoff2018adapting}.  Summaries in SDS datasets such as
CNN/DailyMail \cite{hermann2015teaching}, are on average 30~tokens
long.  In contrast, query focused summaries can be as long as~250
tokens.  We sidestep this problem by adopting a
\textit{retrieval}-based solution.  Specifically, we first build a
database with all summaries in the original dataset.  For each sample
$(d_i, s_i)$, we query the database with summary~$s_i$.  We retrieve
$N_i-1$ other summaries $\mathcal S_i$ with the bigram hashing and
TF-IDF matching method described in \newcite{chen2017opendomain}.
Then, we fetch their corresponding articles $\mathcal D_i$, and form
the $i$th cluster as:
\begin{gather}
    \mathcal D_i^\ast = \{d_i\} \bigcup \mathcal D_i\\
    \hat{s}_i^\ast = \concat(s_i,, s_{i,1}, \ldots, s_{i,N_i}), s_{i,n} \in \mathcal{S}_i
\end{gather}
where $\mathcal D_i^\ast$ are the source documents, and
$\hat{s}_i^\ast$ is a potentially redundant summary of them.
We set $N_i$ to minimize the length difference between
$\hat{s}_i^\ast$ and our summary length requirement (e.g.,~250
tokens).  To obtain the final summary ${s}_i^\ast$, we eliminate
redundancy by selecting sentences from the start of~$\hat{s}_i^\ast$,
skipping sentences that have high cosine similarity 
with those which have already been selected.



\paragraph{Summarization Input} 
In generic MDS, the input to the summarization model is a long
sequence, i.e.,~documents within a cluster are concatenated together
and sentences in each document follow their original order
\cite{fabbri2019multi}.  In QFS, information about absolute (document)
position is lost after evidence ranking. As a result, there is a
discrepancy between training and testing for our generation
model.  To mitigate this, we collect all sentences across documents
for each training sample and rank them in descending order according
to their \mbox{ROUGE-2} score against the reference summary. The
pretrained language model is fine-tuned against this evidence-ranked
list of sentences.
During inference, when \emph{actual} queries are available, we instead use
the top sentences ranked by our query model as input to
summary generation.

\paragraph{Query Guidance} 
Given that summarization input essentially consists of sentences that
are highly relevant to the query, an obvious question concerns the
usefulness of explicitly modeling the query during generation.  We
thus instantiate two conditional language models.  For a
\textit{query-guided} summarizer $p_{\phi}(S|D,Q;\phi)$, we prepend
\texttt{UMRS}$_{S}$ to the selected evidence during training and
\texttt{UMR}$_{Q}$ at inference.  While for a \textit{query-agnostic}
summarizer $p_{\phi}(S|D;\phi)$, we only consider the selected
evidence as input to our summarizer and this setting is identical to generic
MDS.
\paragraph{Length Control} 
QFS tasks usually require summaries of a fixed length budget (e.g, 250
words), whereas summary length is bound to be variable in the training
data.
Inspired by \newcite{lengthControl}, we quantize summary length into
discrete bins.
We augment each training instance with this information, i.e., we
prepend a length token (e.g.,~\texttt{[230]}) to document sentences.
At inference, we inform the model of the summary budget by prepending
the expected length token (e.g.,~\texttt{[250]}) to the sentences
selected by the evidence ranker (see Figure~\ref{fig:frame}(b)).



\section{Experimental Setup}
\paragraph{Datasets} We performed experiments on the DUC 2005-2007 QFS
benchmarks and
TD-QFS \cite{baumel2016topic}. 
DUC benchmarks contain long query
narratives 
while TD-QFS focuses on medical texts with short keyword queries.
Statistics for both datasets are given in Table~\ref{tab:duc_stats}.
We used DUC 2005 as a development
set to optimize hyperparameters and select abstractive models, and
evaluated performance on the other three datasets.

We used Multi-News \cite{fabbri2019multi} and CNN/DailyMail
\cite{hermann2015teaching} as our generic summarization datasets to
train \textsc{Marge} (for evidence ranking) and to fine-tune
\textsc{UniLMv2} (for summary generation).  Data statistics are shown
in Table~\ref{tab:train_stats}.  To create the training and
development sets for optimizing \textsc{Marge}, we sampled sentences
from each dataset.  Specifically, we took the first and last
20~sentences from each cluster in Multi-News and  the first and last three sentences from each article in CNN/DailyMail.
For fine-tuning \textsc{UniLMv2}, we used the original Multi-News and
the synthetic multi-document version of CNN/DailyMail described in
Section \ref{sec:gen}.

\begin{table}[t]
\centering
\setlength{\tabcolsep}{3pt}
\begin{tabular}{@{}l@{~}r@{~}r@{~}r@{~}r@{}}
  \thickhline
Dataset &  2005 &2006  & 2007 & TD-QFS\\
  \thickhline
 Domain  & Cross &Cross &Cross & Medical\\
Query Narrative  &Long&Long&Long & Short\\
\#Clusters & 	50	&    50	  &   45 & 4\\
\#Queries/Cluster 
& 1 &	   1	 &1 & 10\\
\#Documents/Cluster 
& 32 &	   25	 &    25 & 185\\
\#Summaries/Query &	4-9	    & 4	  &   4 & 3\\
\#Words/Summary & 250 & 250 & 250  & 250\\
\thickhline
\end{tabular}
\caption{\label{tab:duc_stats}Multi-Document QFS dataset statistics.}
\end{table}

\begin{table}[t]
\centering
\setlength{\tabcolsep}{3pt}
\begin{tabular}{lrr}  \thickhline
Query Modeling & Multi-News & CNN/DM\\  \thickhline
\#Sentence/Doc & 20 & 3\\
\#Train &1,615,508 &1,719,210\\
\#Validation &200,824 &80,052\\
\#Words/Proxy Query 
& 111.7 & 26.0\\
\#Masks/Proxy Query 
& 35.6 & 8.1\\ \thickhline
\multicolumn{3}{c}{} \\\thickhline
 Summary Generation & Multi-News & CNN/DM\\  \thickhline
\#Clusters 
&44,972 &287,227 \\
\#Documents/Cluster & 2.8 & 4.1
\\
\#Words/Summary & 257.2 & 261.3
\\
\thickhline
\end{tabular}
\caption{\label{tab:train_stats}
  Training data for query modeling and summary generation.
  CNN/DM statistics for summary generation refer to synthetic MDS dataset proposed in this work (based on CNN/DM).}
\end{table}

\paragraph{Implementation Details} 

We used the publicly released BERT
model\footnote{https://github.com/huggingface/pytorch-transformers}
and fine-tuned it for ROUGE regression with a learning rate of
$3\times 10^{-5}$ and a batch size of 128 for 3 epochs on 8 GPUs (GTX
2080 Ti).  We trained two summarization models on CNN/DailyMail and
Multi-News, respectively, with the same hardware.
For both models, we set the maximum input length to 768, and
fine-tuned the publicly released \textsc{UniLMv2}
model\footnote{https://github.com/microsoft/unilm} with a learning
rate of $7\times 10^{-5}$ and a batch size of 16 for 40,000 steps with
gradient accumulation every 4 steps.  During decoding, we used beam
search with beam size $5$ and Trigram Blocking \cite{paulus2018deep}
to reduce redundancy.  The cosine similarity threshold for redundancy
removal was set to~0.6 and summary length was discretized to
10~bins. The $\lambda$~parameter for label smoothing was set to
0.15. We set~$\gamma$, the parameter which modulates the proportion of
information slots to reveal during masking, to~0 (see Appendix for
detailed analysis of $\gamma$ and its effect on model performance).

\section{Results}
\label{sec:results}

Our experiments evaluate both components of the proposed approach,
namely query modeling and summary generation. We assess the evidence
ranker and the effectiveness of the unified masking. We also compare
our summaries against competitive abstractive and extractive systems
using automatic and human-based evaluation.

\begin{table}[t]
\centering
\begin{tabular}{@{}l@{}c@{}c@{}c@{}c@{}c@{}c@{}}
\thickhline
\multirow{2}*{{Models}}  & \multicolumn{2}{@{~}c@{~}}{DUC 2006} &
\multicolumn{2}{@{~}c@{~}}{DUC 2007} & \multicolumn{2}{@{}c@{}}{TD-QFS}\\
 & {\small R@10} & {\small R@30}  & {\small R@10} & {\small R@30} &
 {\small R@10~} & {\small R@30} \\
\thickhline
{\sc Oracle}
&\hspace*{.8ex}6.7 &16.2 
&\hspace*{.8ex}8.4 &19.1
&{17.2} &{35.6} \\
{\sc TermFreq}
&\hspace*{.8ex}7.2 &15.1 
&\hspace*{.8ex}8.5 &18.5 
&{14.2} &{25.9} \\
{\sc  BertQa}
&\hspace*{.8ex}8.5 & 16.3 
& 10.2 & 20.2 
& \hspace*{.8ex}9.8 & 21.9 
\\
{\sc  BertMrc}
&\hspace*{.8ex}8.2 & 16.6 
&\hspace*{.8ex}9.0  & 19.2 
&\hspace*{.8ex}8.1  & 16.4 \\
\hline
{\sc  Marge-Mn}
&\bf{11.1}  &\bf{20.2} 
&\bf{13.8}	&\bf{25.3} 
&11.2 	&21.6 	\\
\hspace{.4cm}$+${\sc Expand} & --- & --- & --- & --- &\bf{18.1}  & \bf{32.9}\\
{\sc  Marge-Cd}
&\hspace*{.8ex}9.1   &17.4	
&11.1    &22.1  
&10.0 	&18.7 	\\
\hspace{.4cm}$+${\sc Expand} & --- & --- & --- & --- & 17.2 & 27.7 \\
\thickhline
\end{tabular}
\caption{\label{tab:rank} 
  Retrieval performance of evidence rankers. 
  R@$k$ is  ROUGE-2 recall  against   top~$k$
  sentences. \textsc{Marge} models are trained on Multi-News
  (\textsc{Mn}) and CNN/DailyMail (\textsc{Cd}) datasets. 
}
\end{table}


\subsection{Query Modeling}
\label{sec:auto_eval_rank}

\paragraph{Evaluation Metrics} 
We evaluate query modeling with \textit{retrieval} and
\textit{summarization} metrics.  For the former evaluation, we follow
\newcite{liu2019hierarchical}, concatenate the top $k$~ranked sentences, and
calculate recall against gold summaries. We additionally propose to
evaluate model output as if it were an extractive summary, to better
assess coverage and informativeness. We thus take the top sentences
subject to a budget of 250~tokens, and remove redundancy by selecting sentences from the top and 
skipping sentences that have high cosine similarity (e.g., $\geq 0.6$)
with selected ones.
  We use ROUGE F1 to evaluate the resulting
summaries so that \textit{precision} is also taken into account.

\paragraph{Results} 
We compare \textsc{Marge} against Term Frequency, a simple but
effective retrieval method that performs particularly well on DUC
datasets \cite{katragadda2009query}. 
We also compare to two semantic
matching models used for extractive QFS \cite{xu2020query}:
\textsc{BertQA} which is trained on the joint set of WikiQA
\cite{yang2015wikiqa} and TrecQA \cite{yao2013answer}, and
\textsc{BertMrc} which is fine-tuned on SQuAD 2.0
\cite{rajpurkar2018know}.
\textsc{Oracle} uses reference summaries as queries to retrieve summary sentences.
For \textit{summarization} evaluation, we report upper bound
performance (\textsc{Gold}) which we estimated by comparing a
(randomly selected) reference summary against the remaining three
reference summaries.  In addition, we compare to \textsc{Lead} which
returns all lead sentences of the most recent document (up to 250
words) and \textsc{LexRank} \cite{erkan2004lexrank}, a widely-used
unsupervised method based on Markov random walks on
sentence-similarity graphs.\footnote{To examine ranking performance,
  we exclude multi-stage frameworks like \citet{xu2020query} that
 rerank the evidence with additional modules (e.g., centrality).}

We summarize ranking and summarization results in
Tables~\ref{tab:rank} and~\ref{tab:ext_main}.
  As we can see,
  despite learning from weak signals, i.e., proxy queries and proxy
  answers, \textsc{Marge} outperforms the strongest baseline,
  \textsc{BertQa}, under both evaluation tasks.  Without recourse to
  any question/answer annotations or dataset-specific retrieval
  methods, our model provides more informative input to the downstream
  generation task. As anticipated, query expansion ($+$\textsc{Expand}) gives
  a big boost on \mbox{TD-QFS} (which has short queries) leading to better
  coverage.

\begin{table}[t]
\tabcolsep=0.15cm
\bgroup
\def\arraystretch{1.0}
\centerline{
\begin{tabular}{@{}l@{}c@{}c@{}c@{}c@{}}
\thickhline
\multirow{1}*{{Models}} 
& \multicolumn{1}{c}{{DUC 2006}} 
& \multicolumn{1}{c}{{DUC 2007}}
& \multicolumn{1}{c}{{TD-QFS}}\\\thickhline
{\sc  Gold} &  17.0 &   19.1  &---\\
{\sc  Oracle} &  14.8 & 16.0 &23.0\\
{\sc  Lead} & 10.4 & 11.3 &10.4 \\
\hline
{\sc TermFreq}	&12.6	&14.2	&12.0\\
{\sc  LexRank}
 & 11.4  & 12.7  &12.2\\
{\sc  BertQa} &13.9	&14.9 &16.1\\
{\sc  BertMrc} &13.6 &14.3  &13.2 \\
\hline
{\sc  Marge-Mn} &	\bf{14.5}	&	\bf{16.6} &15.9 \\
\hspace{.4cm}$+${\sc Expand} & --- & --- & \bf{23.0} \\
{\sc  Marge-Cd} & 13.9&	15.8 &16.9 \\
\hspace{.4cm}$+${\sc Expand} & --- & --- & 22.7  \\
\thickhline
\end{tabular}
}
\egroup
\caption{\label{tab:ext_main} 
  Performance of evidence rankers on extractive QFS.
  We report the F1 score of R-SU4 (for the full set of ROUGE results,
  see Appendix).}
\end{table}

\begin{table}[t]
\tabcolsep=0.1cm
\bgroup
\def\arraystretch{1.0}
\centerline{
\begin{tabular}{lccc}
\thickhline
\multirow{1}*{{Models}}  & \multicolumn{1}{c}{{DUC 2006}} 
& \multicolumn{1}{c}{{DUC 2007}}
 & \multicolumn{1}{c}{{TD-QFS}}\\
\thickhline
{\sc  Marge-Mn} &	{14.5}	&	{16.6} &	{23.0}\\
{\hspace*{2.2ex}$-$Verb} &$\downarrow$0.5    &$\downarrow$0.3 &$\downarrow$2.8\\
{\hspace*{2.2ex}$-$Mask}  &$\downarrow$0.8  &$\downarrow$1.2 &$\downarrow$1.5\\
{\hspace*{2.2ex}$-$Query} &$\downarrow$2.9 &$\downarrow$2.9 &\hspace*{1.1ex}$\downarrow$12.6\\
{\hspace*{2.2ex}$-$OpenIE} &$\downarrow$0.9 & $\downarrow$1.1 & $\downarrow$2.1\\
\thickhline
\end{tabular}
}
\egroup
\caption{\label{tab:ext_ablation} 
  Ablation results on training data (absolute performance decrease denoted by $\downarrow$).}
\end{table}


\paragraph{Ablation Studies}
 Table~\ref{tab:ext_ablation} shows the outcome of various 
 ablation studies which assess  the effectiveness of
masking and how to best instantiate it. 
Specifically,
$-$Verb additionally treats verbs as information slots for sampling
and masking; $-$Mask removes masking entirely so that the whole
summary is revealed; $-$Query removes the proxy query (at training
time) and the actual query (at inference time); this is to investigate
whether our model simply learns to judge sentence salience based on
its own features, instead of performing semantic matching with the
given query; $-$OpenIE removes the dependency on Open IE and chooses
words to mask at random. Specifically, we randomly mask 15\% words in
summaries as in \textsc{Bert} \cite{devlin2019bert} and merge adjacent
\texttt{[MASK]} tokens. Performance drops in all cases, especially
when queries are removed, underscoring the effectiveness of the
proposed representation and training framework.

\subsection{Abstractive Summarization}
\label{sec:auto_eval_summ}

\paragraph{Automatic Evaluation}
Table~\ref{tab:abs_main} compares our model, which we call
\textsc{MargeSum}, against existing QFS systems. These include
\textsc{PQSum-Wsl} \cite{laskar2020wsl} a supervised abstractive
system which represents the state of the art on DUC benchmarks.  It
first extracts relevant sentences for each document with a QA model,
it then replaces some of these with reference summary sentences via a
paraphrase model, and uses them to further fine-tune \textsc{BertSum}
\cite{bertsum}. In its supervised incarnation, two years' DUC datasets
are used for training and one for testing. \textsc{QuerySum}
\cite{xu2020query} is state-of-the-art extractive system which adopts
a coarse-to-fine process for salience estimation.

The second block compares our model with two \textit{distantly
  supervised} approaches. \textsc{Bart-Caq} \cite{su2020caire} uses an
ensembled QA model to extract answer evidence, and fine-tuned
\textsc{Bart} \cite{bart} to iteratively generate summaries from
paragraphs.  \textsc{PQSum} \cite{laskar2020wsl}, uses fine-tuned
\textsc{BertSum} to generate summaries for each document in a cluster,
and a QA model to rank summary sentences against the
query. Table~\ref{tab:sys_comparison} compares these models and our
own in terms of their training requirements.

\begin{table}[t]
\tabcolsep=0.17cm
\bgroup
\def\arraystretch{1.0}
\centerline{
\begin{tabular}{@{}l@{~}c@{~}c@{~}c@{}}
\thickhline
\multirow{1}*{{Models}} 
& \multicolumn{1}{@{}c@{~}}{{DUC 2006}} 
& \multicolumn{1}{@{~}c@{~}}{{DUC 2007}}
& \multicolumn{1}{@{~}c@{}}{{TD-QFS}}\\ \thickhline
{\sc PQSum-Wsl}$^\dagger$
&  16.5 & 17.7 & --- \\
{\sc QuerySum}$^\ast$
& 15.3 & 16.8 & 20.7 \\
\hline
{\sc Bart-Caq}   &12.9 &14.4 & --- \\
{\sc PQSum}  & 14.8  & 16.0 & --- \\
\hline
{\sc  UniLM-Mn} 	&11.8	&12.3 &12.9\\
{\sc  UniLM-Cd}     &13.6	&14.9	&16.7
\\
\hline
{\sc  MargeSum-Mn}
	&14.3	&16.5 &16.5
\\
{\sc  MargeSum-Cd}
	&\bf{15.1}	& \bf{16.9}
&\bf{20.9}\\
\thickhline
\end{tabular}
}
\egroup
\caption{\label{tab:abs_main} 
   Abstractive summarization models with \mbox{R-SU4} (full set of results
   in Appendix);    $\ast$/$\dagger$: extractive/supervised method.
}
\end{table}

\begin{table}[t]
\centering
\setlength{\tabcolsep}{3pt}
\begin{tabular}{@{}l@{}@{}c@{~}c@{~}c@{~}c@{}}
  \thickhline
Models & QA & PI & GS & QFS\\
  \hline
{\sc Bart-Caq} \cite{su2020caire} & \cmark & \xmark & \cmark & \xmark \\
{\sc PQSum} \cite{laskar2020wsl} & \cmark & \xmark & \cmark & \xmark\\
{\sc PQSum-Wsl} \cite{laskar2020wsl} & \cmark & \cmark & \cmark & \cmark\\
{\sc UniLM} \cite{bao2020unilmv2} & \xmark & \xmark & \cmark & \xmark\\
\hline
{\sc MargeSum} & \xmark & \xmark & \cmark & \xmark\\
\thickhline
\end{tabular}
\vspace*{-.2cm}
\caption{\label{tab:sys_comparison}
  Training requirements for existing QFS models (QA, PI, GS, and QFS
  stand for question answering, paraphrase identification, generic
  summarization and query focused summarization).} 
\end{table}


The third block presents the performance of \textsc{UniLM} fine-tuned
on Multi-News and CNN/DailyMail following the standard setting in
\newcite{bao2020unilmv2}.  It uses no query guidance or length
control.  Documents are concatenated as input for training. During
testing, sentences are selected with \textsc{Marge} but ordered
according to their original document position.  The last block shows
two variants of \textsc{MargeSum}, optimized on Multi-News and a
synthetic training set built from CNN/DailyMail.  Both take as input
sentences selected with \textsc{Marge-Mn} during inference.


As we can see, without requiring expensive QA data (see
Table~\ref{tab:sys_comparison}), \textsc{MargeSum-Cd} outperforms
existing distantly supervised approaches.  Its performance on DUC is
on par with one of the strongest extractive systems, while on TD-QFS
it is superior across metrics.  Also note that \textsc{Marge} trained
on synthetic MDS data outperforms \textsc{MargeSum-Mn}.  Compared to
Multi-News, synthetic summaries cover more topics and are less
redundant, which is suited to  QFS  where there are usually
multiple sub-queries to answer.
\begin{table}[t]
\tabcolsep=0.1cm
\bgroup
\def\arraystretch{1.0}
\centerline{
\begin{tabular}{@{}l@{}c@{}c@{}c@{}c@{}c@{}c@{}}
\thickhline
\multirow{1}*{{Models}} 
& \multicolumn{1}{@{}c@{~~}}{{DUC 2006}}
& \multicolumn{1}{@{}c@{~~}}{{DUC 2007}}
& \multicolumn{1}{@{}c@{}}{{TD-QFS}}\\\thickhline
{\sc Marge-Cd}
&{15.1} &{16.9} &{20.9}\\
{\hspace*{1.5ex} {\sc BertQa}} 
&$\downarrow$1.0
& $\downarrow$2.2   
&$\downarrow$6.1\\ 
{\hspace*{2ex}$-$Rank}  
&$\downarrow$1.7
&$\downarrow$3.1
&$\downarrow$1.3
\\
{\hspace*{2ex}$-$Length} 
&$\downarrow$0.1
&$\downarrow$0.5
&$\downarrow$0.2
\\
{\hspace*{2ex}$-$Query} 
&$\downarrow$0.5
&$\downarrow$0.3
&$\downarrow$0.4\\
\thickhline
\end{tabular}
}
\egroup
\caption{\label{tab:abs_ablation} Ablations for
  \textsc{MargeSum} trained on CNN/Daily Mail (performance decrease
  denoted by~$\downarrow$).  
}
\end{table}

\paragraph{Ablation Studies} 

 Table~\ref{tab:abs_ablation} presents the results of several ablation
 studies on \textsc{MargeSum-Cd}.  Replacing the input to the
 summarization component with sentences selected by \textsc{BertQa} \cite{xu2020query} significantly decreases performance, demonstrating
 that sentences selected by \textsc{Marge} are useful to downstream
 abstractive summarization.  Removing evidence ranking altogether
 ($-$Rank) leads to a large performance drop; this is expected since
 sentence position information from the original documents does not
 transfer well to QFS settings.  Removing length control ($-$Length)
 also hurts performance as does the removal of query guidance
 ($-$Query) at inference time.

\paragraph{Human Evaluation} 
\begin{table}[t]
\centering
\begin{tabular}{llll} \thickhline
    DUC & {Rel} & {Suc} & {Coh}\\
  \thickhline
   {\sc Gold}&	3.05	&3.29&	3.35\\
   {\sc PQSum-Wsl} &\textbf{2.95}	&\textbf{3.27}	&2.93$^{\dagger\circ}$\\
{\sc QuerySum} &2.79	&3.13	&2.94$^{\dagger\circ}$\\ 
{\sc UniLM-Cd} &2.43$^{\dagger\circ}$	&3.09	&3.27 \\ 
{\sc MargeSum-Cd} &2.91	&3.25	&\textbf{3.30}\\ 
\hline
  \multicolumn{4}{c}{}\\\hline
 TD-QFS & {Rel} & {Suc} & {Coh}\\\hline
{\sc Gold} &4.70	&4.23	&4.60 \\
{\sc QuerySum} &4.32	&3.90$^{\circ}$	&3.80$^{\dagger\circ}$\\ 
{\sc UniLM-Cd} &3.63$^{\dagger\circ}$	&\textbf{4.12}	&4.28\\ 
{\sc MargeSum-Cd}& \textbf{4.55}	&4.02	&\textbf{4.37}\\
\thickhline
\end{tabular}
\caption{\label{tab:human_eval} Human evaluation results on DUC
  (above) and TD-QFS (below): average \textbf{Rel}evance,
  \textbf{Suc}cinctness, \textbf{Coh}erence ratings; 
  $\dagger$: sig different
  from \textsc{MargeSum-Cd}; 
  $^\circ$: sig different from Gold (at
  \mbox{$p < 0.05$}, using a pairwise t-test). 
  }
\end{table}


We also evaluated model summaries in a judgment elicitation study via
Amazon Mechanical Turk. Native English speakers (self-reported) were
asked to rate query-summary pairs on two dimensions: Succinctness
(does the summary avoid unnecessary detail and redundant information?)
and Coherence (does the summary make logical sense?).  The ratings
were obtained using a fivepoint Likert scale.  In addition,
participants were asked to assess the Relevance of the summary to the
query.  Crowdworkers read a summary and for each sentence decided
whether it is relevant (i.e., it provides an answer to the query),
irrelevant (i.e., it does not answer the query), and partially
relevant (i.e., it is not clear it directly answers the
query). Relevant sentences were awarded a score of 5, partially
relevant ones a score of 2.5, and 0 otherwise. Sentence scores were
averaged to obtain a relevance score for the whole summary.

Participants assessed summaries created by
\mbox{\textsc{PQSum-Wsl}}, the state-of-the-art abstractive system,
\textsc{QuerySum}, a state-of-the-art extractive system,
\textsc{UniLM-Cd}, and \textsc{MargeSum-Cd}.\footnote{We are
grateful to Md Tahmid Rahman Laskar for providing us with the output
of their \textsc{PQSum-Wsl} system.  We include \textsc{PQSum-Wsl}
only for human evaluation on DUC since it was not evaluated on TD-QFS
\cite{laskar2020wsl} and system output is not available.}
We also randomly selected  \textsc{Gold} standard summaries to
include as an upper bound.  We sampled 20 query-cluster pairs from DUC
(2006, 2007; 10 from each set), and 20 pairs from TD-QFS (5 from each
cluster) and collected three responses per  pair.

Table \ref{tab:human_eval} shows the human ratings for each system (we
provide examples of summary output in Appendix C).  Participants
perceive \textsc{MargeSum-Cd} on par with \textsc{PQSum-Wsl} in terms
of query relevance and summary succinctness, while significantly
better than \textsc{PQSum-Wsl} and \textsc{QuerySum} in terms of
coherence.  In fact, participants find summaries \textsc{PQSum-Wsl}
summaries as incoherent as those created by the extractive
\textsc{QuerySum}; this is probably due to the fact that
\textsc{PQSum-Wsl} first generates an abstractive summary for each
document and then re-ranks the generated sentences. Therefore, final
summary sentences are less related to each other.
Summaries from our system are also considered significantly more relevant than  \textsc{UniLM-Cd}.
Compared to \textsc{PQSum-Wsl}, although \textsc{UniLM-Cd} is not
good at producing relevant content, it maintains relatively higher
coherence, demonstrating the effectiveness of training abstractive
systems with synthetic data from SDS and generating long summaries at
once. 
\section{Conclusions}
\label{sec:conclusions}
In this work we proposed an abstractive framework for query focused
summarization. We provided a unified mask representation for summaries
and queries, which enables summaries to serve as proxy queries for
model training. As a result, a query model can be trained with generic
summarization data without relying on additional question-answering
resources.
Experimental
results across datasets show that the proposed system yields
state-of-the-art performance despite the weakly supervised setting,
and produces more relevant and coherent summaries compared to existing
approaches. In the future, we would like to push this low-resource
approach even further and attempt to generate abstractive summaries
without access to any summarization datasets.

\subsection*{Acknowledgments}
The authors would like to thank the anonymous reviewers for their
valuable feedback. We acknowledge the financial support of the
European Research Council (Lapata; award number 681760).  This
research is based upon work supported in part by the Office of the
Director of National Intelligence (ODNI), Intelligence Advanced
Research Projects Activity (IARPA), via contract FA8650-
17-C-9118. The views and conclusions contained herein are those of the
authors and should not be interpreted as necessarily representing the
official policies or endorsements, either expressed or implied, of the
ODNI, IARPA, or the U.S. Government. The U.S. Government is authorized
to reproduce and distribute reprints for Governmental purposes
notwithstanding any copyright annotation therein.

\bibliographystyle{acl_natbib}
\bibliography{tacl.bib}

\newpage
\appendix

\section{Evidence Ranking Results}
\begin{table}[h]
\tabcolsep=0.1cm
\bgroup
\def\arraystretch{1.0}
\centerline{
\begin{tabular}{lccc}
\hline
{Models}
& {DUC 2006}
& {DUC 2007}
& {TD-QFS}\\
\thickhline
{\sc Oracle}
&22.7 &26.2 &{44.6} \\
{\sc TermFreq}
&20.8 &25.2 &{34.0} \\
{\sc  BertQa}
& 22.1 & 26.1 & 29.1 \\
{\sc  BertMrc}
& 22.3 & 25.2 &23.2
\\
\hline
{\sc  Marge-Mn}
&\bf{25.9} &\bf{31.8} &29.4\\
\hspace{.4cm}$+${\sc Expand} 
& --- & --- &\bf{39.1}\\
{\sc  Marge-Cd}
&\hspace*{.8ex}23.3 &28.8 &26.2\\
\hspace{.4cm}$+${\sc Expand} 
& --- & --- &26.2
\\
\thickhline
\end{tabular}
}
\egroup
\caption{\label{tab:rank2} 
  Performance of evidence rankers on top retrieval.
  We report the ROUGE~2 recall score for the concatenation of the top 50 retrieved sentences. 
}
\end{table}

We show in the paper the top~$k$ retrieval performance of different
models when $k \in \{10,30\}$.  In some cases, when top sentences are
relatively short, the maximum input length to \textsc{UniLM} (which is
set to 768) allows for more than~30 sentences to be selected.
Therefore, in Table~\ref{tab:rank}, we further show the top~$k$
retrieval performance of evidence rankers with larger $k$, set
to~$k=50$.  Results show that our model outperforms strong baseline
systems, and we conclude that it consistently provides high quality
content, under varied budgets ($k \in \{10, 30, 50\}$), to the
downstream abstractive summarization task.

We report the full set of ROUGE results for evidence rankers on
extractive summarization in the main paper in Table~\ref{tab:ext_main}.  

\begin{table*}[h]
\bgroup
\def\arraystretch{1.0}
\centerline{
\begin{tabular}{lccc ccc ccc}
\thickhline
\multirow{2}*{Models} 
& \multicolumn{3}{c}{{DUC 2006}} 
& \multicolumn{3}{c}{{DUC 2007}}
& \multicolumn{3}{c}{{TD-QFS}}\\
~ & {R-1} & {R-2} & {R-SU4} 
& {R-1} & {R-2} & {R-SU4} 
& {R-1} & {R-2} & {R-SU4}\\
\thickhline
{\sc  Gold} & 45.7 &11.2 &  17.0 & 47.9 &14.1 &  19.1  &-&-&-\\
{\sc  Oracle} & 40.6 &\hspace*{.8ex}9.1 &  14.8 & 41.8 &10.4 & 16.0 & 44.9 &18.9 &23.0\\
{\sc  Lead} &  32.1 &\hspace*{.8ex}5.3 & 10.4 & 33.4 & \hspace*{1.2ex}6.5 & 11.3 
&33.5 &\hspace*{1.2ex}5.2 &10.4
\\
\hline
{\sc TermFreq}
&36.5	&\hspace*{.8ex}7.0	&12.6
&38.5	&\hspace*{1.2ex}9.0	&14.2
&35.7	&\hspace*{1.2ex}6.5	&12.0\\
{\sc  LexRank}
& 34.2 & \hspace*{.8ex}6.4 & 11.4 & 35.8 & \hspace*{1.2ex}7.7 & 12.7 
&35.3 &\hspace*{1.2ex}7.6 &12.2\\
{\sc  BertQa}
&38.6 &\hspace*{.8ex}8.4	&13.9	
&39.8	&10.0	&14.9
&39.5	&10.5	&16.1
\\
{\sc  BertMrc}
&\bf{39.6}	&\hspace*{0.8ex}7.8	&13.6
&39.9	&\hspace*{1.2ex}8.9	&14.3
&36.6	&\hspace*{1.2ex}8.4	&13.2
\\
\hline
{\sc  Marge-Mn}
&39.0	&\hspace*{.8ex}\bf{9.3}&	\bf{14.5}	&\bf{41.6}	&\bf{11.6}&	\bf{16.6}
&38.8	&10.5	&15.9 \\
\hspace{.4cm}$+${\sc Expand} 
& --- & --- & --- 
& --- & --- & --- 
& \bf{45.9} &\bf{18.8} &\bf{23.0}\\
{\sc  Marge-Cd}
&38.4   &\hspace*{.8ex}8.6&	13.9
&40.7   &10.8&	15.8
&40.1 &11.6 &16.9\\
\hspace{.4cm}$+${\sc Expand} 
& --- & --- & --- 
& --- & --- & --- 
& \bf{45.9} &18.3&22.7
\\
\thickhline
\end{tabular}
}
\egroup
\caption{\label{tab:ext_main2} 
  Performance of evidence rankers on extractive QFS.
  R-1, R-2 and R-SU4 stand for the F1 score of ROUGE~1, 2, and SU4, respectively.
}
\end{table*}
\begin{table*}[t]
\tabcolsep=0.17cm
\bgroup
\def\arraystretch{1.0}
\centerline{
\begin{tabular}{lccc ccc ccc}
\thickhline
\multirow{2}*{{Models}} 
& \multicolumn{3}{c}{{DUC 2006}} 
& \multicolumn{3}{c}{{DUC 2007}}
& \multicolumn{3}{c}{{TD-QFS}}\\
~ & {R-1} & {R-2} & {R-SU4} 
& {R-1} & {R-2} & {R-SU4} 
& {R-1} & {R-2} & {R-SU4}\\
\thickhline
{\sc PQSum-Wsl}$^\dagger$ \cite{laskar2020wsl}
& 43.5 & 10.8 & 16.5
& 44.7 & 12.4 & 17.7
& - & - & - \\
{\sc QuerySum}$^\ast$ \cite{xu2020query}
& 41.6 & \hspace*{.8ex}9.5 & 15.3
& 43.3 & 11.6 & 16.8
& 44.3 & 16.1 & 20.7
\\
\hline
{\sc Bart-Caq} \cite{su2020caire} 
&38.3 &\hspace*{.8ex}7.7 &12.9
&40.5 &\hspace*{.8ex}9.2 &14.4
& - & - & - \\
{\sc PQSum} \cite{laskar2020wsl}
&\bf{40.9} &\hspace*{.8ex}9.4 & 14.8 
& 42.2 & 10.8 & 16.0
& - & - & - \\
\hline
{\sc  UniLM-Mn} 
&34.6	&\hspace*{.8ex}6.7	&11.8
&35.5	&\hspace*{.8ex}7.6	&12.3
&36.2   &\hspace*{.8ex}8.1	&12.9
\\
{\sc  UniLM-Cd} 
&37.6	&\hspace*{.8ex}8.3	    &13.6	
&39.6	&10.1	&14.9	
&40.1	&11.8	&16.7
\\
{\sc  MargeSum-Mn}
&39.1	&\hspace*{.8ex}9.1	&14.3	
&42.1	&11.7	&16.5
&40.8	&11.6	&16.5
\\
{\sc  MargeSum-Cd}
&{40.2}&\hspace*{.8ex}\bf{9.7}	&\bf{15.1}	
&\bf{42.5}	&\bf{12.0}&	    \bf{16.9}
&\bf{45.5}	&\bf{16.6}	&\bf{20.9}\\
\thickhline
\end{tabular}
}
\egroup
\caption{\label{tab:abs_main_full} 
  Performance of abstractive summarization systems. 
  R-1, R-2 and R-SU4 stand for the F1 score of ROUGE~1, 2, and SU4, respectively. 
  $\ast$/$\dagger$: extractive/supervised method.
}
\end{table*}
\begin{table}[h]
\begin{tabular}{c}
\vspace{2cm}
\end{tabular}
\end{table}
\section{The Effect of Reveal Ratio}
\begin{figure}[h]
	\includegraphics[width=7.5cm]{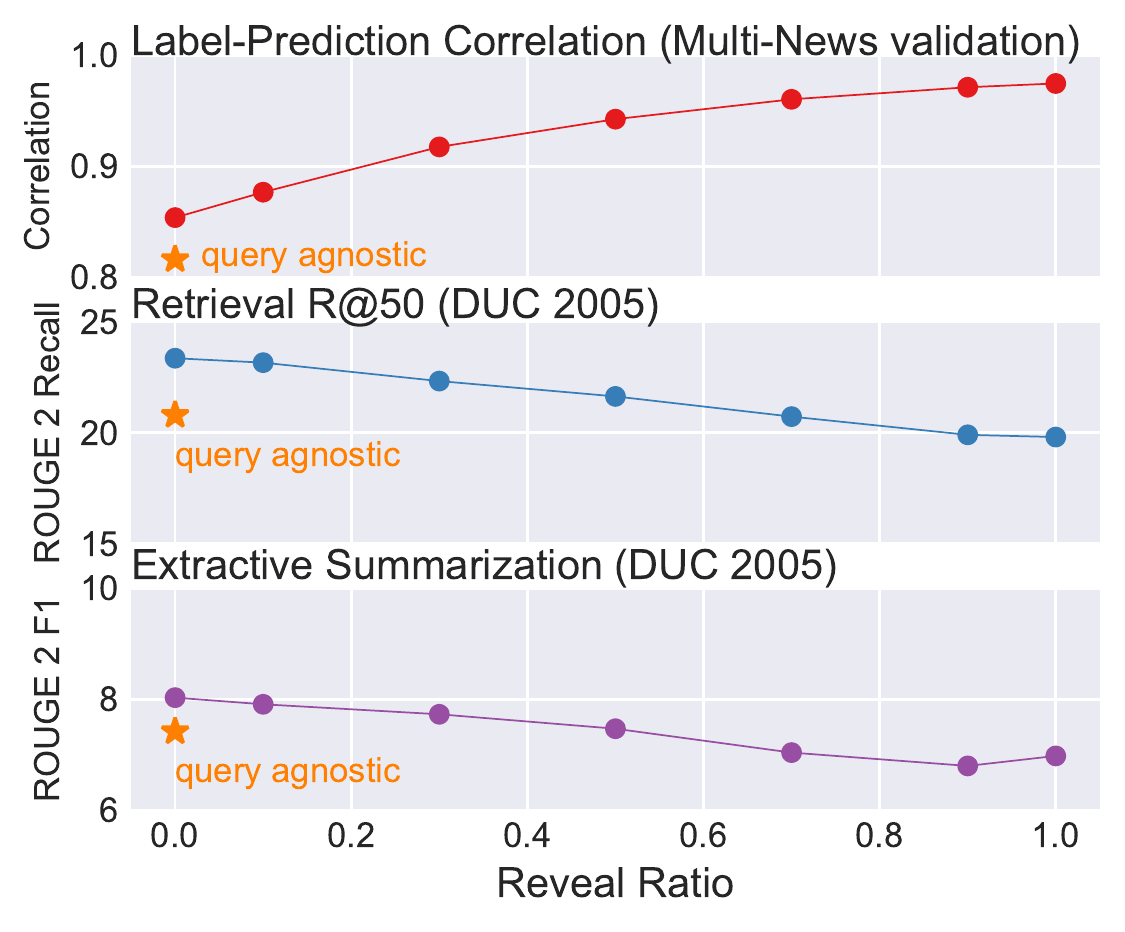}
	\caption{\label{fig:mask} Model performance when reveal ratio
          $\gamma$ is varied. Correlation refers to the average of
          Pearson's $r$ correlation.  The star marker denotes
          query-agnostic performance where all query tokens are
          masked, including information slots.  }
\end{figure}

We show how the mask reveal ratio $\gamma$ affects model performance
in Figure~\ref{fig:mask}.  As we can see, performance on the ROUGE
regression task improves as $\gamma$ increases; this is not
surprising, the task becomes easier when fewer tokens are masked; when
$\gamma=1.0$, simply counting lexical overlap can solve the task
perfectly.  However, model performance on the QFS development set (DUC
2005) shows the opposite trend: actual queries \textit{seek}
information, instead of providing all the information needed.
Therefore, the model is required to perform \textit{semantic matching}
\cite{guo2016deep} to accurately estimate evidence scores.  Based on
our empirical results, a simple but effective strategy is to mask all
information slots (i.e., potential arguments) and reveal the rest of
the words (including verbs) in the summary to construct proxy queries
for training.

\section{Abstractive Summarization Results}
We report the full set of ROUGE results for abstractive summarization
models in Table~\ref{tab:abs_main_full}. We also show an examples of
system outputs in Table~\ref{tab:example}.

\begin{table*}[t]
\small
\bgroup
\def\arraystretch{1.0}
\centering
\begin{tabular}{@{~}p{15.5cm}@{~}}\hline
\textbf{Query}: 
Steroid use among female athletes. Discuss the prevalence of steroid use among female athletes over the years. Include information regarding trends, side effects and consequences of such use.
\\\hline
\textbf{\textsc{Gold}}: 
\textcolor{ggreen}{Increasing numbers of female athletes use anabolic steroids.
Among female college athletes, an estimated 5\% in swimming, 6\% in basketball and 10\% in track and field have used steroids.
The fastest growing group of users is adolescent females.
New steroids are often marketed as dietary supplements and girls are unaware of what they are taking.
A young tennis player tested positive for steroids after taking an "all natural" supplement later discovered to contain a variation on the banned steroid nandrolone.
A former Olympic track coach believes at least 40\% of the U.S. women's team in Seoul had used steroids at some time.}
\textcolor{gblue}{Because of their naturally low testosterone levels, steroids have a more dramatic effect on women, boosting their levels up to 10 times.
The health effects can be dramatic, including liver damage and tumors, elevated cholesterol, heart attacks, strokes, stunted growth in adolescents, infertility, uncontrollably violent behavior, chronic depression, deeper voices, excessive facial hair, and acne.
Many East German female athletes were given steroids, often unbeknownst to them or their parents and suffered serious side-effects.}
  \textcolor{gred}{Athletes who have tested positive for steroids include a Bulgarian triple jumper, a Romanian hammer thrower, a Russian runner, a Dominican high, a Jamaican sprinter, a Spanish pole vaulter, a German marathon runner, two Moroccan athletes, and two Chinese Taipei weightlifters.
A number of female Chinese athletes have tested positive, including several swimmers and rowers.}
  \textcolor{gyellow}{International organizations impose bans of between two months and life on athletes found guilty of using illegal steroids.}
 \\\hline
\textbf{\textsc{UniLM-Cd}}: 
\textcolor{gred}{At least 23 Chinese athletes, mostly women, have tested positive for steroid use since 1990.}
Such side-effects of anabolic steroids have been generally known since 1977.
\textcolor{gred}{Some of the athletes were as young as 12 when they started taking the steroids.}
EPO is one of at least 20 steroids prohibited by the International Olympic Committee.
The drug is legal and sold over-the-counter, but is banned by the National Olympic Committee, the National Football League and the National Collegiate Athletic Association.
The National College Athletic Association (ACA) has banned EPO for life.
EPO has been linked to a number of health problems, including liver damage and liver cancer.
The ACA says EPO's use in the U.S.is safe, effective and safe to use.
The National Institute of Drug Abuse (NIDA) is investigating the use of EPO in the United States.
The NIDA says it has no evidence to support the claim that EPO can cause liver damage.
The drug was approved for use by the U.S. Olympic committee in 1996.
The National Association of College Athletes (NAA) is calling for the drug to be banned from all U.S. sports.
The NAA says it's time for the U.S. Olympic Committee (IOC) to ban EPO.
\textcolor{gred}{The drug has been used by more than 200,000 athletes worldwide since it was approved in 1996.}
 \\\hline
\textbf{\textsc{PQSum}}: 
\textcolor{ggreen}{About 175,000 teen age girls in the United States have reported using anabolic steroids at least once within a year of the time surveyed a rise of 100 percent since 1991.}
\textcolor{gred}{In 1996 Samantha Reeves became the first female tennis player to have tested positive for steroids.}
The rise of steroid use in female athletes has led to a rise in women's steroid use.
Over the years the number of female steroid related women in the U.S has increased
There is no mistaking the dangers of steroids for women with documentation provided by trainers and coaches from the former east german sports empire.
\textcolor{gyellow}{In 1996 women were banned from the olympics for a positive test of methandienone.}
\textcolor{gblue}{The women involved suffered serious side effects from the drugs and that they knew of the potential danger.}
She was found to have steroid stanozolol in her system.
\textcolor{gred}{In the last month Cuban sprinter Merlene Ottey withdrew from the world.championships after testing positive for nandrolone.}
\textcolor{gred}{Some were as young as 12 when they started taking the anabolic steroid pills.}
The drug use has been generally known since 1977.
\textcolor{gred}{Pospelova would be the seventh athlete to test positive at the games.}
\textcolor{ggreen}{Such side effects of anabolic steroids are generally known as with all probability linked to doping.
An over the counter supplement called andro raises testosterone and estrogen above normal levels and could be dangerous according to a harvard study by major league baseball and its players union.}
In 1996.
\\\hline
\textbf{\textsc{MargeSum-Cd}}:
\textcolor{ggreen}{Penn State professor Charles Yesalis estimates the use of steroids among female athletes at 5 percent in swimming, 6 percent in basketball and 10 percent in track and field, a rise of 100 percent since 1991.
The national institutes of drug abuse says 175,000 teenage girls in the united states have reported taking anabolic steroids at least once within a year of the time surveyed.
The national institute on drug abuse provides information regarding trends, side effects and consequences of such use.}
\textcolor{gyellow}{Two Moroccan female athletes have been stripped of gold and bronze medals for using a muscle-building steroid in the first reported cases of doping at the Arab games for using the steroid nandrolone,} \textcolor{gblue}{a steroid that has been linked to liver cancer, heart disease and uncontrollable aggressiveness.}
  \textcolor{gred}{Two medical experts testifying in the doping trial of a former east german sports} \textcolor{gblue}{doctor say the female swimmers they examined showed health damage linked to performance-enhancing drugs, including liver damage and excessive facial hair.}
The study, published in Wednesday's Journal of the American Medical Association, is the first to conclude that high doses of the steroids can elevate testosterone levels and that the hormone can be used as a performance-enhancing steroid, such as epitestosterone, as a marker the testosterone is 6 to 1 in the male sex hormone and 5 to 1 for the female steroid hormone epitestoterone - a metabolite that is used as an indicator of testosterone use - the female sex hormone.
 \\\hline
\end{tabular}
\caption{\label{tab:example}
System outputs for cluster D0602C in DUC 2006.  
The gold summary answers the query covering four main aspects (denoted
  with different colors): (1) \textcolor{ggreen}{trend;}
  (2) \textcolor{gblue}{side-effects;} (3)
  \textcolor{gyellow}{consequences of such use;} (4)
  \textcolor{gred}{historical cases.} 
  Both outputs from \textsc{MargeSum-Cd} and \textsc{PQSum} have a good coverage of the main query focuses.
   Compared to \textsc{PQSum}, \textsc{MargeSum-Cd} produces a more coherent summary for the given query narrative with a more natural topic flow.}
\egroup
\end{table*}

\section{Datasets and Evaluation Package}
Multi-News and CNN/Daily Mail are used to train the query model and
abstractive summarization model described in this work, and they can
be downloaded from \url{https://github.com/Alex-Fabbri/Multi-News} and
\url{https://github.com/abisee/cnn-dailymail}, respectively.

For evaluation purposes, the TD-QFS dataset is publicly available at \url{https://www.cs.bgu.ac.il/~talbau/TD-QFS/dataset.html}. 
DUC 2005-2007 benchmarks can be requested from NIST: \url{https://www-nlpir.nist.gov/projects/duc/data.html}. 

We computed ROUGE scores  with \texttt{pyrouge}, a Python wrapper for the ROUGE summarization evaluation package: \url{https://github.com/bheinzerling/pyrouge}.
\end{document}